\documentclass{llncs}

\usepackage{booktabs}
\usepackage{caption} 
\usepackage{amsmath}
\usepackage{graphicx}
\usepackage{pdfpages}
\graphicspath{ {images/} }
\usepackage{enumitem}
\usepackage{tabularx}
\usepackage{tabulary}
\usepackage{mytitlesec}
\usepackage{fixltx2e}

\titlespacing\section{0pt}{10pt plus 4pt minus 1pt}{2pt plus 2pt minus 1pt}
\titlespacing\subsection{0pt}{12pt plus 4pt minus 2pt}{0pt plus 2pt minus 2pt}
\titlespacing\subsubsection{0pt}{12pt plus 4pt minus 2pt}{0pt plus 2pt minus 2pt}

\begin{document}
\title{Unsupervised Open Relation Extraction}

%
%
\author{Hady Elsahar\inst{1}, Elena Demidova \inst{2}, Simon Gottschalk \inst{2},
Christophe Gravier\inst{1}, Frederique Laforest\inst{1}
}

\authorrunning{Elsahar et al.} 

%
\institute{
Univ Lyon, UJM-Saint-Etienne, CNRS, Laboratoire Hubert Curien, France 
\email{(hady.elsahar,christophe.gravier,frederique.laforest)@univ-st-etienne.fr}
\and
L3S Research Center, Leibniz Universität Hannover, Hannover, Germany
\email{(demidova, gottschalk)@L3S.de}
}

\maketitle
\begin{abstract}
%
%
%
We explore methods to extract relations between named entities from free text in an unsupervised setting. 
%
In addition to standard feature extraction, we develop a novel method to re-weight word embeddings.
%
We alleviate the problem of features sparsity using an individual feature reduction.
Our approach exhibits a significant improvement by $5.8\%$
over the state-of-the-art relation clustering
 scoring a F1-score of $0.416$ on the NYT-FB dataset.


\end{abstract}
\section{Introduction}

Relation extraction (RE) is the task of identification and classification of relations between named entities (such as persons, locations or organizations) in free text. 
%
RE is of utmost practical interest for various fields including event detection, knowledge base construction and question answering. 
Fig. \ref{fig:input} illustrates a typical RE task.
For the first two sentences, RE should identify the semantic relation type \textit{birth place} between the named entity pairs
regardless of the surface pattern used to express the relation such as 
\textit{hometown is} or \textit{was born in}. RE should also distinguish it from the  
album production relation between the same named entities in the third sentence.
\begin{figure}[h]
\small
    \begin{enumerate}[topsep=0pt,partopsep=0pt]
        \item \textit{\textbf{David Bowie}}'s \underline{hometown is} \textit{\textbf{London}}, United Kingdom.
        \item \textit{\textbf{Axel Rose}}, also known as "William Bruce", 
        \underline{was born in} \textit{\textbf{Lafayette}}, Indiana.
        \item \textit{\textbf{David Bowie}} \underline{produced his first album in} \textit{\textbf{London}}, United Kingdom. 
    \end{enumerate}
    \caption{Sentences containing textual \underline{relations} between \textit{\textbf{named entities}}. 
    }\label{fig:input}
\end{figure}

%
%
Distant supervision techniques for RE~\cite{distant_re,aug} have proven to be very efficient in solving that problem. However, distant supervision is limited to a fixed set of relations in a given knowledge base, which hinders its adaptation to new domains. 
%
%
Unsupervised approaches~\cite{yao2011,marcheggiani2016} can potentially overcome these limitations 
by applying purely unsupervised methods enabling extraction of open relations (relations 
unknown in the knowledge base in advance).
In this paper, we propose an unsupervised approach to extract and cluster open relations between named entities from free text
by re-weighting word embeddings and using the types of named entities as additional features. 

\section{Proposed Method}
\begin{figure}
 \centering 
 \includegraphics[width=0.9\textwidth]{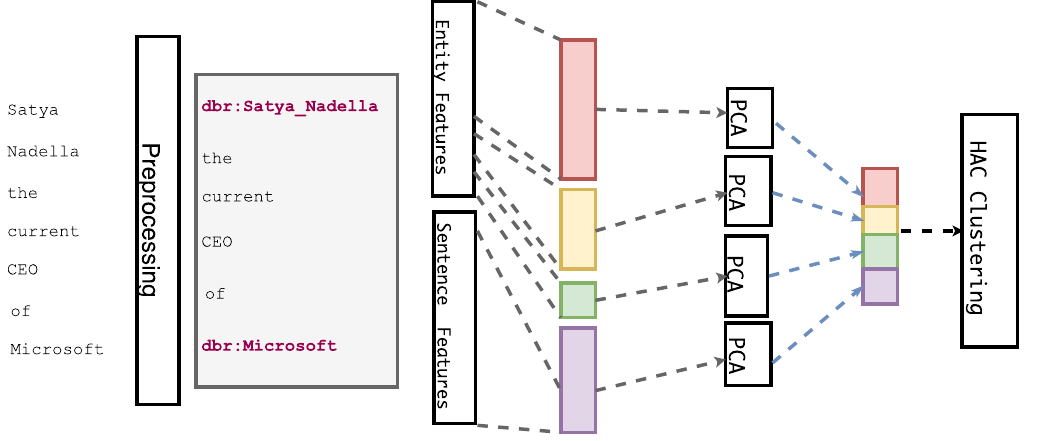}
 \caption{System overview}\label{fig:systemoverview}

\end{figure}

Our system builds sentence representations based on the types of the involved named entities, %
and the terms forming the relations. For the latter, we use pre-trained word embeddings after re-weighting them according to the dependency path between the named entities. 
These representations are clustered so that different representations of the semantically equivalent relations are mapped to the same cluster. 
Fig.~\ref{fig:systemoverview} presents an overview of our system for unsupervised open relation extraction, consisting of four stages: preprocessing, 
feature extraction, sparse feature reduction and relation clustering
described in the following.
%


\noindent\textbf{Preprocessing}
%
%
%
For each sentence in the dataset, we extract named entities using DBpedia Spotlight 
and consider all sentences containing at least two entities. For this set of sentences, the Stanford CoreNLP dependency parser is utilized to extract the lexicalized dependency path between each pair of named entities.\\

\noindent\textbf{Feature Extraction}
For each sentence, our method outputs a 
vector representation of the textual relation between each named entity pair.
Features include word embeddings, dependency paths between named entities, and named entity types.
%
%
Word embeddings provide an estimation of the semantic similarity between terms using vector proximity. 
%
%
Sentence representations are typically built by averaging word vectors. 
However, not all words in a sentence equally contribute to the expression of the relation between two named entities. 
%
%
Therefore we develop a novel method to re-weight the pre-trained word embeddings. 
Terms that appear within the lexicalized dependency path between the two named entities are given a higher weight.
Intuitively, shorter dependency paths are more likely to represent true relationships between the named entities. 
The vector representation $s(W,D)$ of each sentence is calculated through the following function:
\[s(W,D) = \sum_{w_i \in W} f(w_{i},W,D) \cdot v(w_{i}), \quad
    f(w_{i},W,D)= 
\begin{cases}
    \frac{C_{in} \cdot |W| }{|D|},& \text{if } w_{i} \epsilon D \\
    C_{out},              & \text{otherwise}
\end{cases},
\]


\noindent where $W=\{w_1,...,w_n\}$ is the set of terms in the sentence, $D \subset W$ is the set of terms in the lexicalized dependency path between the named entities in the sentence, and $v(w_{i})$ is the pre-trained word embedding vector for $w_i$. $C_{in} \geq 1$ and $C_{out}$ are constant values experimentally set to $1.85$ and $0.02$. 
We use Glove\footnote{http://nlp.stanford.edu/projects/glove/}
word embeddings of size 100.
As a baseline, we compare these representations with standard sentence representations features such as: TF-IDF, 
the sum of word embeddings, and the sum of IDF re-weighted word embeddings~\cite{sentsimilarity}.
Intuitively, relations can connect entities of certain types. 
For example, a birth place relation connects a person and a location, although other
relations between person and location are possible. 
Therefore, for each named entity, we use its DBpedia types and Stanford NER tags as features.\\ 

\noindent\textbf{Sparse Feature Reduction}
Some of the features are more sparse than the others; concatenating them for each relation skews the clustering. 
In supervised relation extraction, this is not an issue as any learning algorithm is expected to do feature selection automatically using the training data. 
In unsupervised relation extraction there is no training data, hence we devise a novel strategy in order to circumvent the sparse features bias. 
Individual feature reduction of the sparse features is applied before merging them with the rest of the feature vectors.
For feature reduction, we use Principal Component Analysis (PCA)~\cite{pca}.\\

\noindent\textbf{Relation Clustering}
We use Hierarchical Agglomerative Clustering (HAC) to cluster the feature representations of each relation, with Ward's~\cite{ward1963hierarchical} linkage criteria\footnote{accessing the clustering output by HAC at rank $k$ giving $k$ clusters}, which yields slightly better results than the k-means clustering algorithm. 

\section{Evaluation}
To evaluate our system, we use the NYT-FB dataset~\cite{marcheggiani2016}. This dataset contains approximately 1.8M sentences divided into 80\%-20\% test-validation splits 
and aligned automatically to the statements (triples) from Freebase. 
The alignment between sentences and the properties of the Freebase triples in this dataset is considered as the gold standard for the relation clustering algorithm.\\
We use the validation split to tune the parameters for re-weighting word vectors and the PCA algorithm, and the test set for evaluating relation discovery methods.
We compare our method using the best identified feature combination with the state-of-the-art models for unsupervised Relation Discovery, namely the variational autoencoders model \cite{marcheggiani2016}
and two other systems, Rel-LDA \cite{yao2011}, and Hierarchical Agglomerative Clustering (HAC) baseline with standard features \cite{yao2012}. 
To make our results comparable we set the number of relations to induce (number of clusters $k$) to 100, following the SOA systems. \\
Table~\ref{table:feat_eval} shows the performance of the clustering algorithm by relying only on sentence representations as features. Results demonstrate that our method of word embeddings re-weighted by the dependency path shows a significant improvement over other traditional sentence representations.
Table~\ref{table:eval} shows the performance when the dependency re-weighted word embeddings are merged with the rest of the proposed features and applying individual feature reduction. Our method outperforms the state-of-the-art relation discovery algorithm scoring a pairwise F1 score of 41.6\%.
    \begin{table}[h]
        \begin{minipage}{0.5\textwidth}
            \centering
                \begin{tabularx}{\textwidth}{ll}\toprule
                    Feature                                 & F\textsubscript{1}      \\   \midrule
                    TF-IDF                                  & 12.2              \\    
                   Word-Emb.                         & 7.4               \\     
                    IDF-Emb.                          & 10.3              \\    
                    Dependency Re-Weighted Emb. & \textbf{19.5} \\ \bottomrule
                \end{tabularx}
            \caption{\label{table:feat_eval} Comparison between different 
            features for clustering.}
        \end{minipage}
        \begin{minipage}{0.05\textwidth}
        ~
        \end{minipage}
        \begin{minipage}{0.45\textwidth}
            \centering
                 \begin{tabularx}{\textwidth}{llll}\toprule
                    Var. Auto\-encoder & Rel-LDA & HAC & Our  \\   \midrule    
                    35.8                & 29.6      & 28.3       & \textbf{41.6}     \\ \bottomrule
                \end{tabularx}
            \caption{\label{table:eval}  Pairwise F\textsubscript{1} (\%)  scores of different models on the test set of the NYT-FB dataset.}
        \end{minipage}
    \end{table}

%
%

%

\section{Conclusion}

In this paper, we proposed an approach for unsupervised relation extraction from free text. 
%
Our approach is based on a novel method of re-weighting word vectors according to the dependency 
parse tree of the sentence. 
As additional features, we use the types of named entities involved in the relations.
A final HAC clustering is applied to the sentence representations so that similar representation of a relation are mapped to the same cluster. %
Our evaluation results demonstrate that our method outperforms the state-of-the-art relation clustering method by 5.8\% pairwise F1 score.
The code for feature building and dimensionality reduction is publicly 
available\footnote{https://github.com/hadyelsahar/relation-discovery-2-entities.git}.

\footnotesize{
 \section*{Acknowledgements}    
 This work was partially funded by H2020-MSCA-ITN-2014 WDAqua (64279), ALEXANDRIA (ERC 339233) and Data4UrbanMobility (BMBF).
 }

\footnotesize{
\bibliography{eswc2017}
\bibliographystyle{splncs03}
}

\end{document}